\pdfoutput=1

\documentclass[11pt]{article}

\usepackage[]{emnlp2021}

\usepackage{times}
\usepackage{latexsym}

\usepackage[T1]{fontenc}

\usepackage[utf8]{inputenc}

\usepackage{microtype}
\usepackage{enumitem}
\usepackage{setspace}
\usepackage{graphicx}
\usepackage{subcaption}
\usepackage{amsmath, amsthm, amssymb}
\usepackage{bm}
\usepackage{algorithm,algorithmic}
\usepackage{listings}
\usepackage{booktabs,siunitx}
\usepackage{bm}
\usepackage{bbm}
\usepackage{multirow}
\usepackage{threeparttable}
\usepackage{stfloats}
\usepackage{float}

\title{Enhanced Seq2Seq Autoencoder via Contrastive Learning for Abstractive Text Summarization}

\author{Chujie Zheng{$^{1,*}$}, Kunpeng Zhang{$^{2}$}, Harry Jiannan Wang{$^{1}$}, Ling Fan{$^{3,4}$}, Zhe Wang{$^{4}$} \\
        {$^{1}$}University of Delaware \\ {$^{2}$}University of Maryland, College Park \\ {$^{3}$}Tongji University, {$^{4}$}Tezign \\
        \texttt{chz@udel.edu}, \texttt{kpzhang@umd.edu}, \texttt{hjwang@udel.edu},\\\texttt{lfan@tongji.edu.cn}, \texttt{w@tezign.com}}

\date{}

\begin{document}
\maketitle
\begin{abstract}
In this paper, we present a denoising sequence-to-sequence (seq2seq) autoencoder via contrastive learning for abstractive text summarization. Our model adopts a standard Transformer-based architecture with a multi-layer bi-directional encoder and an auto-regressive decoder. To enhance its denoising ability, we incorporate self-supervised contrastive learning along with various sentence-level document augmentation. These two components, seq2seq autoencoder and contrastive learning, are jointly trained through fine-tuning, which improves the performance of text summarization with regard to ROUGE scores and human evaluation. We conduct experiments on two datasets and demonstrate that our model outperforms many existing benchmarks and even achieves comparable performance to the state-of-the-art abstractive systems trained with more complex architecture and extensive computation resources.
\end{abstract}

\setlength{\abovedisplayskip}{3pt}
\setlength{\belowdisplayskip}{3pt}
\setlist[itemize]{noitemsep, topsep=0pt}

\section{Introduction}

Text summarization aims to produce an accurate text snippet to capture the key information. Existing methods are either extractive or abstractive. Extractive methods select sentences from the document and the abstractive methods generate sentences based on the input document as a summary. With the advancement of natural language processing (NLP) research, especially in the area of large-scale pre-trained language models \cite{devlin2019bert, peters2018deep, radford2019language, liu2019roberta} in recent years, abstractive summarization has become a popular research topic and made significant progress. In most of existing abstractive  summarization models, such as BART \cite{lewis2020bart}, PEGASUS \cite{zhang2020PEGASUS} and ProphetNet \cite{qi2020prophetnet}, all adopt Transformer-based architecture \cite{vaswani2017attention}. They are usually first pre-trained in an unsupervised manner with a large amount of corpus and then fine-tuned on a specific dataset for supervised downstream applications. These models have shown superiority on various text understanding tasks, especially for generating abstractive summaries. 
\begin{figure}[t]
\centering
\includegraphics[width=0.5\textwidth]{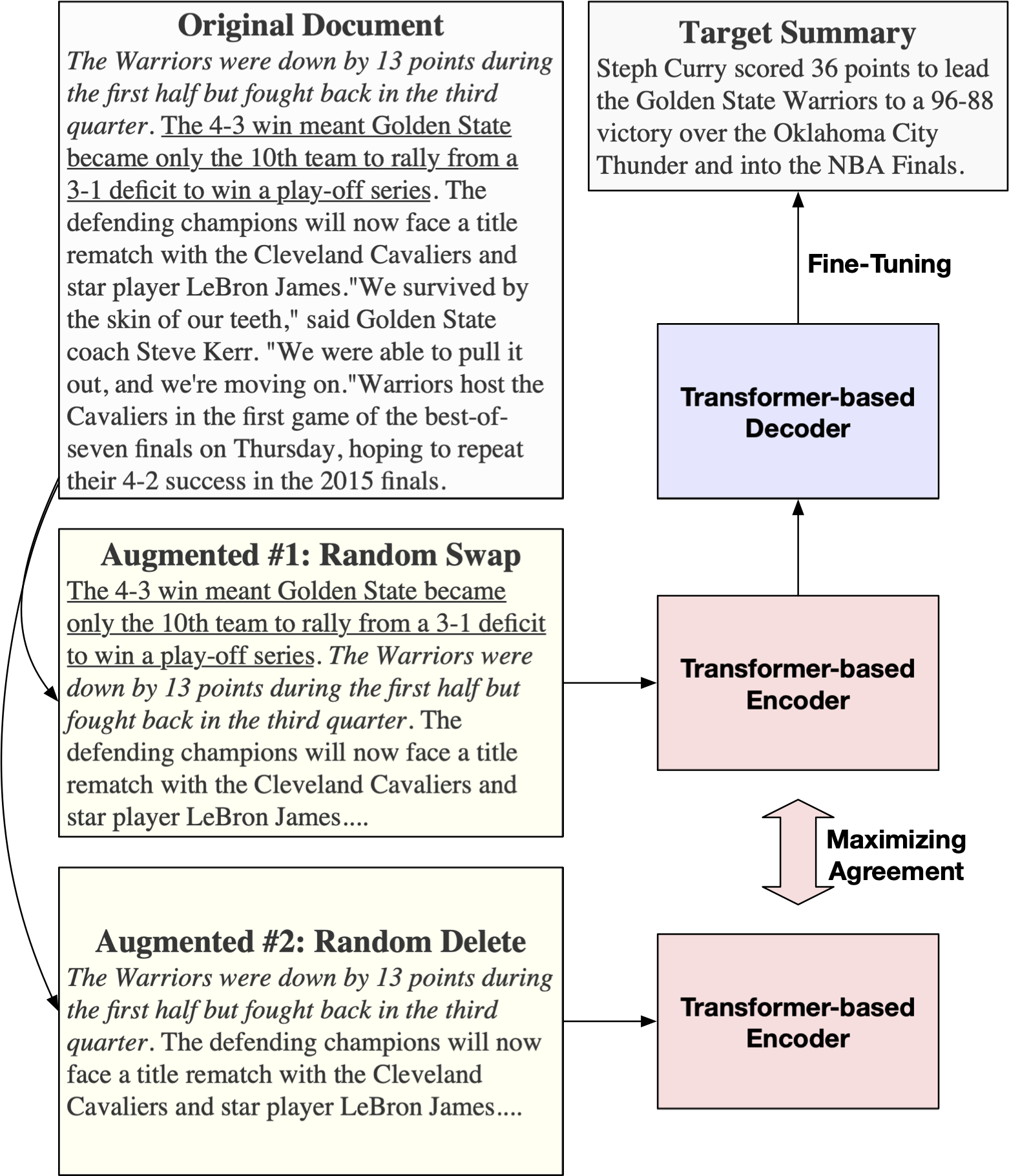}

\caption{An illustration example of ESACL.} 
\label{fig::figure1-framework}
\end{figure}

Despite impressive performance on standard benchmarks, these deep networks are often brittle when deployed in real-world systems \cite{goel2021robustness}. The primary reason lies in that they are not robust to various noises, such as data corruption \cite{belinkov2018synthetic}, distribution shift \cite{hendrycks2020many} or harmful data manipulation \cite{jia2017adversarial}. In addition, they may also heavily rely on spurious patterns for prediction \cite{mccoy2019right}. As demonstrated in prior studies, the seq2seq model plays a critical role in many downstream applications. Thus, we expect to enable its denoising capability when developing such a seq2seq model in NLP tasks. 
Furthermore, many prior studies in language understanding find that the global semantics may significantly be neglected by Transformer-based models \cite{fang2020cert}. Because self-attention in these models is usually applied to learn and predict word-level characteristics during pre-training. The sentence embeddings aggregated from word embeddings learned by existing pre-trained language models have been found not be able to effectively and sufficiently capture the semantics among sentences \cite{li2020sentence}. This can lead to poor performance for subsequent tasks, e.g., abstractive summarization. The reason is that summarization requires wide-coverage natural language understanding going beyond the meaning of individual words and sentences \cite{liu2019text}. Therefore, to build a denoising seq2seq model, state-of-the-art (SOTA) approaches like BART \cite{lewis2020bart} and MARGE \cite{lewis2020pre} developed new objectives for pre-training. BART is trained by first corrupting documents at a word level and then optimizing a reconstruction loss between the generated output and the original document. MARGE learns the model by self-supervising the reconstruction of target text where it first retrieves a set of related texts and then maximize the likelihood of generating the original documents based on selected texts. All these seq2seq-based approaches are inspirational and emphasize the ability of denoising and modeling global semantics.

In this study, we propose a new framework \textbf{ESACL}, \underline{E}nhanced \underline{S}eq2Seq \underline{A}utoencoder via \underline{C}ontrastive \underline{L}earning, to improve the denoising ability of the seq2seq model and increase the model flexibility by achieving our goal through fine-tuning. Unlike most existing methods that design denoising objectives in pre-training, ESACL optimizes the model in the fine-tuning phase which requires less computation resources and significantly saves training time. Specifically, ESACL leverages self-supervised contrastive learning \cite{chen2020simple, he2020momentum} and integrates it into a standard seq2seq autoencoder framework. Overall, it involves two stages: (1) sentence-level document augmentation, and (2) joint learning framework of seq2seq autoencoder and contrastive learning with an overall objective based on a fine-tuning loss and a self-supervised contrastive loss. Regarding the seq2seq autoencoder, ESACL uses a similar architecture to BART, which is a standard transformer-based model with a multi-layer bi-directional encoder and left-to-right decoder. As shown in Figure \ref{fig::figure1-framework}, ESACL performs document augmentation to create two instances, and designs a unique framework underlying the seq2seq model: it not only uses the output from the decoder for fine-tuning but also tries to maximize agreement of the output from the encoder between two augmented instances.

A key step in contrastive learning is data augmentation. Various augmentation strategies have been developed in many NLP tasks at the word level, such as inserting a new word or swapping two tokens.
To capture high-level semantics and the structural information of the entire document, we perform data augmentation at the sentence level. In this study, we implement several combinations of data augmentation and our experiment results show that (i) the model performance can be improved with sentence-level augmentation; (ii) the summarization performance with different data augmentation strategies does not vary much; (iii) the augmentation that largely interrupts the structure of the document should be avoided. 

To sum up, ESACL proposes a new way of denoising a seq2seq model via fine-tuning for abstractive summarization. It presents a new scheme for summarization which incorporates self-supervised contrastive learning into a seq2seq framework to improve the model flexibility. The major contributions of this study are as follows:
\begin{itemize}[leftmargin=*]
    \itemsep0em 
	\item We propose ESACL, a new abstractive text summarization framework that jointly trains a seq2seq autoencoder with contrastive learning through fine-tuning.
	\item We evaluate ESACL using two summarization datasets through quantitative measurement, robustness check, and human evaluation. ESACL achieves state-of-the-art performance and has shown better flexibility concerning modeling potential irrelevant noises.
	\item We introduce several sentence-level document augmentation strategies and conduct an ablation study to understand their impact on the performance.
\end{itemize}

\section{Related Work}
\label{section::related-work}

Three lines of research are closely related to our paper: abstractive text summarization, contrastive learning, and data augmentation.

\textbf{Abstractive text summarization} has achieved promising results with the rapid development of deep learning. Neural network-based models \cite{rush2015neural, nallapati-etal-2016-abstractive, chopra2016abstractive, nallapati2017summarunner, zhou2017selective,tan2017abstractive, gehrmann-etal-2018-bottom, zhu2019ncls} enable the framework for generating abstractive summary. Recently, with the success of attention mechanism and Transformer-based \cite{vaswani2017attention} language models, pre-training based methods \cite{devlin2019bert, radford2018improving, radford2019language} have attracted growing attention and achieved state-of-the-art performances in many NLP tasks, and pre-training encoder-decoder Transformers \cite{song2019mass, lewis2020bart, zhang2020PEGASUS, qi2020prophetnet, lewis2020pre} show great successes for the summarization.

\textbf{Contrastive learning} has been recently a resurgence in image analysis and language understanding \cite{khosla2020supervised, chen2020simple, fang2020cert, gunel2020supervised}. Researchers have developed many contrastive learning-based frameworks, including self-supervised framework \cite{fang2020cert} and supervised framework \cite{gunel2020supervised} and apply them to different language understanding tasks, e.g., sentiment analysis \cite{li2020cross} and document clustering \cite{shi2020self}. They mainly use contrastive learning to help models deeply explore the unique characteristics of data while ignoring irrelevant noises, which also motivates the present study.

\begin{figure*}
\centering
\includegraphics[width=1\textwidth]{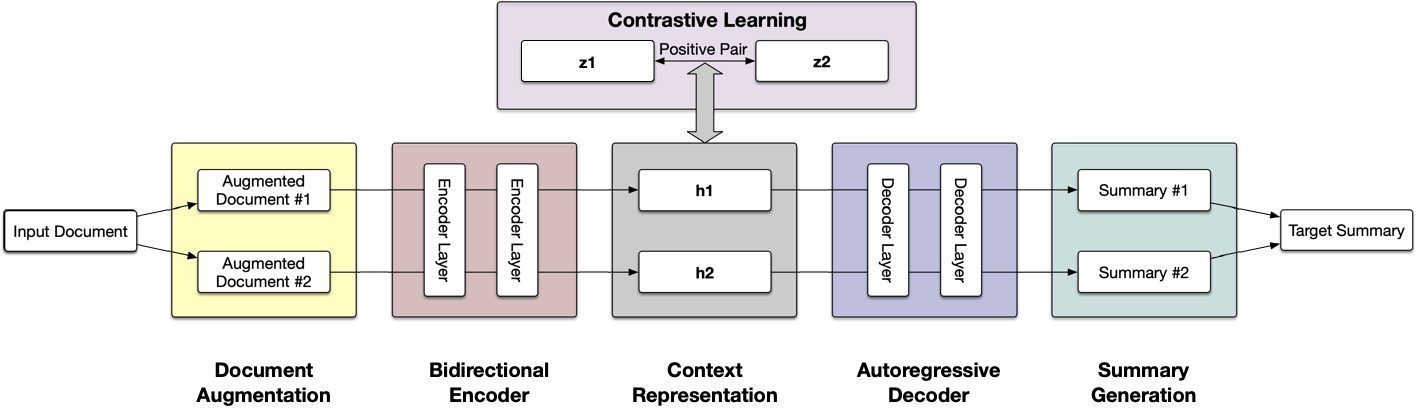}
\caption{The overall architecture of our proposed ESACL.}
\label{fig::model-design}
\end{figure*}

\textbf{Data augmentation} is the key in contrastive learning and has been widely applied in image analysis \cite{wong2016understanding}. Textual data augmentation is different and can be mainly categorized into word-level transformation \cite{ kolomiyets2011model, wang2015s, zhang2015character, qiu2020easyaug} and neural text generation \cite{sennrich2016improving, yu2018qanet}. In our paper, to preserve the global semantics while filtering irrelevant noise for a document, we design several sentence-level augmentation strategies and show their effectiveness in summarization. Based on the experiment results, we believe that developing new alternative augmentations for text summarization has its great merit.

\section{Preliminary}
Automatic text summarization aims at condensing a document to a shorter version while preserving the key information. Let $\textbf{d} = \left\{\textbf{x}_1, \textbf{x}_2, ..., \textbf{x}_N\right\}$ be an input document with $N$ tokens and $\textbf{x}_i$ is the word embedding for the $i$-th token. Given a document $\textbf{d}$, we expect to learn a function $f(\textbf{d})$ that maps $\textbf{d}$ to another sequence of tokens $\textbf{y}=\left\{\textbf{y}_1, \textbf{y}_2, \cdots, \textbf{y}_m\right\}$, where $\textbf{y}$ is the generated summary with $m$ tokens.  $m$ is an unknown apriori and depends on the input sequence and the task-specific requirement.

Such a function $f(\cdot)$ is often implemented by a seq2seq model. The key idea is to represent an input sequence as a low-dimensional vector while preserving the contextual information in the sequence as much as possible, upon which a new task-specific sequence with an arbitrary length can be automatically generated \cite{jurafsky2020speech}. A typical seq2seq model usually consists of three components:
\begin{itemize}[leftmargin=*]
    \itemsep0em 
	\item An \textbf{encoder}, denoted as $f_{\text{encoder}}$ that accepts an input sequence $\textbf{d}$, and generates a corresponding sequence of contextualized representation $\textbf{h}$.
	\item A \textbf{context vector}, $\textbf{c}$ that is a function of $\textbf{h}$ and conveys the essence of the input to the decoder.
	\item And a \textbf{decoder}, $f_{\text{decoder}}$ that uses $\textbf{c}$ to generate an arbitrary length of sequence $\textbf{y}$ based on the task-specific requirement.
\end{itemize}

\section{Our Proposed Model}
In this section, we present our proposed model ESACL, which leverages self-supervised contrastive learning to enhance the denoising ability of a seq2seq framework. Figure \ref{fig::model-design} illustrates the overall architecture of ESACL. For a given input document $\textbf{d}$, ESACL first creates a pair of augmented documents that are expected to associate with the same original target summary. ESACL then generates the latent representation of the augmented documents using the Transformer-based encoder and performs self-supervised contrastive learning to encourage the model to capture potential noises in the document $\textbf{d}$. Finally the optimized latent representation is sent to the Transformer-based decoder to generate the summary. In Section \ref{section::self-supervised-contrastive-learning}, we present our implementation of contrastive learning in ESACL. In Section \ref{section::document-augmentation}, we introduce several sentence-level document augmentation strategies which are the key in contrastive learning. In Section \ref{section::sequence-to-sequence}, we describe the detailed seq2seq architecture of ESACL, in particular how the self-supervised contrastive learning is incorporated and how they are jointly trained via fine-tuning.

\subsection{Document Augmentation}
\label{section::document-augmentation}
Data augmentation has been used to increase the denoising capability of a model. In prior literature as mentioned in Section \ref{section::related-work}, there exists many contrastive learning-based models and applications in NLP. However, most of these methods focus on augmentation at the word level, which might not be suitable for text summarization because the global semantics and even noises at a higher level (e.g., sentence or document) can be easily ignored. In this study, we perform document augmentation at the sentence level. Specifically, given an input document $\textbf{d}$ with a sequence of $k$ sentences, we manipulate the document via various transformations at the sentence level to augment the document. By doing this, ESACL can generate another sequence $\hat{\textbf{d}}$ where main semantics are preserved with some additional noises.

Similar to \citet{qiu2020easyaug}, we design several document augmentation approaches at the sentence level, as follows:
\begin{itemize}[leftmargin=*]
    \itemsep0em 
	\item \textbf{Random Insertion (RI):} randomly pick an existing sentence and insert it into a random position in the input document.
	\item \textbf{Random Swap (RS):} randomly select two sentences and swap their positions.
	\item \textbf{Random Deletion (RD):} randomly delete a sentence from the input document.
	\item \textbf{Document Rotation (DR):} randomly select a sentence and the document is rotated using this selected sentence as the pivot.
\end{itemize}

\subsection{Self-Supervised Contrastive Learning}
\label{section::self-supervised-contrastive-learning}
We introduce self-supervised contrastive learning into ESACL during the fine-tuning process to enhance its noising flexibility. 
ESACL performs document augmentation of the original input document to create positive training pairs. Along with negative pairs (two different documents), ESACL is able to encourage itself to identify if two context vectors learned from the encoder are representing the same original input document. By doing so, ESACL improves the quality of the context vector $\textbf{c}$ during the fine-tuning which can benefit the performance of downstream language generation.

To form positive pairs during training, we perform document augmentation to create two augmented instances for each document in a batch of $K$ training instances $b=\left\{\textbf{d}_1, \textbf{d}_2, ..., \textbf{d}_K \right\}$. Suppose $\textbf{d}_i$ is the original input document, we generate the augmented documents $\hat{\textbf{d}}_{2i-1}=A_1(\textbf{d}_i)$ and $\hat{\textbf{d}}_{2i}=A_2(\textbf{d}_i)$, where $A$ refers to a specific augmentation strategy. Thus, we have $2K$ augmented instances in total for a batch, and we assume $\hat{\textbf{d}}_{2i-1}$ and $\hat{\textbf{d}}_{2i}$ are augmented from the same input document $\textbf{d}_i$. A positive pair is defined if and only if two instances are from the same original input document. Otherwise they are considered as a negative pair. We use the pre-trained encoder $f_{\text{encoder}}(\cdot)$ to obtain the latent representation of each augmented document $\hat{\textbf{d}}$ as $\textbf{h}=f_{\text{encoder}}(\hat{\textbf{d}})$. In our work, we use the final hidden vector corresponding to the first input token as the aggregate representation for the document like prior literature did \cite{devlin2019bert}. ESACL also applies a non-linear projection head $g$ to further understand the deep semantics among latent dimensions. It projects the representation $\textbf{h}$ into another latent space $\textbf{z}=g(\textbf{h})$, which is used to calculate the contrastive loss $l(i,j)$ for the positive pair as Equation \ref{eq::loss}. Here $\mathbbm{1}_{[k \neq i]}$ is 1 when $k \neq i$ and 0 otherwise. $\tau$ is a temperature parameter. $\text{sim}(\cdot, \cdot)$ is a cosine similarity measure.
\begin{equation}
	l(i,j) = -\log \frac{\exp(\text{sim}(\textbf{z}_i, \textbf{z}_j)/\tau)}{\sum_{k=1}^{2K} \mathbbm{1}_{[k \neq i]} \exp(\text{sim}(\textbf{z}_i, \textbf{z}_k)/\tau)}
	\label{eq::loss}
\end{equation}
The loss of contrastive learning in ESACL is:
\begin{equation}
	\mathcal{L}_{\text{cl}} = \frac{1}{2K}\sum_{i=1}^{K}[l(2i-1, 2i)+l(2i, 2i-1)]
	\label{eq::cl-loss}
\end{equation}



\subsection{Sequence-to-sequence Architecture}
\label{section::sequence-to-sequence}
For the abstractive text summarization, we follow the literature and adopt the Transformer-based seq2seq model, which has proven to be effective (see Section \ref{section::related-work}). A natural question arising here is how to leverage the denoising ability of contrastive learning in the seq2seq framework to improve the summarization. To answer this question, we design a combined loss to jointly learn model parameters. For each instance $\textbf{d}_i$, we obtain two augmented instances: $\hat{\textbf{d}}_{2i-1}$ and $\hat{\textbf{d}}_{2i}$ \footnote{Different augmentation strategies can be combined. For example, $\hat{\textbf{d}}_{2i-1}$ is augmented via \textbf{RI} while $\hat{\textbf{d}}_{2i}$ is via \textbf{RS}. }, which are considered as a positive pair for self-supervised contrastive learning. They are also used to generate summaries $\hat{\textbf{y}}_{2i-1}$ and $\hat{\textbf{y}}_{2i}$. The generated summary is compared with the target summary of the original input document for calculating the fine-tuning loss, $\mathcal{L}_{\text{generate}}$, which measures the generation performance. In this study, we define the $\mathcal{L}_{\text{generate}}$ as the cross-entropy loss. We also use the generated positive pair to calculate the contrastive learning loss as we introduced, which measures the noising flexibility of our model. Equation \ref{eq::loss-function} summarizes the overall loss of ESACL as the weighted sum of two losses. A hyper-parameter $\alpha \in [0,1]$ is used to balance the importance of contrastive learning and the summary generation. The overall process of ESACL is summarized in Algorithm \ref{alg::the_alg}.
\begin{equation}
	\mathcal{L} = \alpha \mathcal{L}_{\text{cl}}	+ (1-\alpha)\mathcal{L}_{\text{generate}}
	\label{eq::loss-function}
\end{equation}

\renewcommand{\algorithmicrequire}{\textbf{Input:}}
\renewcommand{\algorithmicensure}{\textbf{Output:}}
\begin{algorithm}

\caption{ESACL training in one epoch}
\label{alg::the_alg}
\begin{algorithmic}[1]
\REQUIRE batch size $K, f_{encoder}, f_{decoder}, g$
\STATE {Pick two augmentation strategies $A_1$, $A_2$}
\FOR{each batch $b \in \left\{1,...,B\right\}$}
\FOR{each document $i \in \left\{1,...,K\right\}$ in $b$} 
\STATE {\color{gray}\# using the first augmentation}
\STATE $\hat{\textbf{d}}_{2i-1}=A_1(\textbf{d}_i)$; \hfill {\color{gray}\# augmented instance}
\STATE $\textbf{h}_{2i-1}=f_{\text{encoder}}(\hat{\textbf{d}}_{2i-1})$; 
\STATE $\textbf{z}_{2i-1}=g(\textbf{h}_{2i-1})$; \hfill {\color{gray}\# projection}
\STATE $\hat{\textbf{y}}_{2i-1}=f_{\text{decoder}}(\textbf{h}_{2i-1})$; \hfill {\color{gray}\# generation}
\STATE {\color{gray}\# using the second augmentation}
\STATE $\hat{\textbf{d}}_{2i}=A_2(\textbf{d}_i)$; \hfill {\color{gray}\# augmented instance}
\STATE $\textbf{h}_{2i}=f_{\text{encoder}}(\hat{\textbf{d}}_{2i})$; 
\STATE $\textbf{z}_{2i}=g(\textbf{h}_{2i})$; \hfill {\color{gray}\# projection}
\STATE $\hat{\textbf{y}}_{2i}=f_{\text{decoder}}(\textbf{h}_{2i})$; \hfill {\color{gray}\# generation}
\ENDFOR
\STATE \textbf{calculate} $\mathcal{L}$ using Equation \ref{eq::loss-function};
\STATE $\theta\leftarrow \underset{\theta}{\arg\min}\mathcal{L}(f_{encoder}, f_{decoder}, g \mid \theta)$;
\ENDFOR
\STATE \textbf{return} the learned $f_{encoder}^*, f_{decoder}^*, g^*$.
\end{algorithmic}
\end{algorithm}

\section{Experiments}
\subsection{Experiment Setting}
We evaluate our model using two popular summarization datasets: the CNN/Daily Mail dataset (CNN/DM) \cite{hermann2015teaching} and the extreme summarization dataset (XSUM) \cite{narayan2018don}. Our experiments are conducted with 3 NVIDIA V100 GPUs. We adopt a 12-layer encoder and a 6-layer decoder with 16 attention heads. We warm-start the model parameter with the distil-BART pre-trained model\footnote{We choose distil-BART provided by HuggingFace. For CNN/DM, we use "sshleifer/distilbart-cnn-12-6". For XSUM, we use "sshleifer/distilbart-xsum-12-6". Appendix \ref{appendix::implementation} records the detailed implementation.} and trains 5 epochs with a batch size of 16\footnote{It takes about 35 hours for 5 epochs on our machine.}. For projection head in contrastive learning, we implement a 2-layer MLP to project the representation to a 128-dimensional latent space.  We use Adam optimizer with a learning rate of $5e-7$.

Given the limited computing resource (e.g., the memory limitation), we need to freeze some layers of the encoder to reduce the number of parameters. The impact of freezing different layers of the encoder will be discussed in the following ablation study (see Section \ref{section::ablation-study}). All results reported below are based on freezing the first 6 layers of the encoder. For the loss calculation, we set $\alpha=0.2$ and $\tau=0.5$. For data augmentation, we choose two augmentation operations, and discuss this hyper-parameter in Section \ref{section::ablation-study}. For the purpose of reproducibility, all codes are publicly available here\footnote{https://github.com/chz816/esacl}. 

\subsection{Experimental Results}
We compare our proposed model with the following cutting-edge summarization models.
\begin{itemize}[leftmargin=*]
    \itemsep0em 
	\item \textbf{Lead-N} uses the first $N$ sentences of the article as its summary.
	\item \textbf{BERTSUM} \cite{liu2019text} proposes a novel document-level encoder based on BERT to generate summary.
	\item \textbf{MATCHSUM} \cite{zhong-etal-2020-extractive} is an extractive summarization approach which formulates the task as a semantic text matching problem.
	\item \textbf{PGNet} \cite{see2017get} is the pointer-generator network, which copies words from the source text and retains the ability to produce novel words. \textbf{PGNet+Cov} is with the coverage mechanism.
	\item \textbf{BART} \cite{lewis2020bart} employs the bidirectional encoder to enhance the sequence understanding and the left-to-right decoder to generate the summary.
	\item \textbf{PEGASUS} \cite{zhang2020PEGASUS} introduces a new pre-train objective to encourage the model generate target sentences, which enables the model to capture global information among sentences.
	\item \textbf{ProphetNet} \cite{qi2020prophetnet} predicts the next $n$ tokens simultaneously based on previous context tokens at each time step. 
\end{itemize}

We adopt ROUGE \cite{lin2004rouge} F1 score as the evaluation metric. We choose ROUGE-1, ROUGE-2, and ROUGE-L for performance measurement, which are the common choices in the literature. We report the performance for all baseline models using the numbers from the original literature.

\textbf{Results on CNN/DM}: Table \ref{table::performance-cnndm} records the performance on CNN/DM. ESACL outperforms most of the baseline models and achieves the highest ROUGE-L score on CNN/DM. Comparing to the SOTA extractive system MATCHSUM, ESACL achieves a higher ROUGE-2 and ROUGE-L score. Comparing to three SOTA abstractive systems, ESACL outperforms ProphetNet and improves the performance of BART by 7.3\% on ROUGE-L. Our model achieves comparable performance with PEGASUS, which is the best-performed SOTA model.

\begin{table}[htbp]
\centering
\begin{threeparttable}
\begin{tabular}{lccc}
\toprule
\multicolumn{1}{c}{\textbf{Model}} & \textbf{RG-1} & \textbf{RG-2} & \textbf{RG-L} \\
\hline
Lead-3 & 40.07 & 17.68 & 36.33 \\
\hline
BERTSUM & 42.13 & 19.60 & 39.18 \\
MATCHSUM & \textbf{44.41} & 20.86 & 40.55 \\
\hline
PGNet & 36.44 & 15.66 & 33.42 \\
PGNet+Cov & 39.53  & 17.28 & 36.38 \\
BART & 44.16 & 21.28 & 40.90 \\
ProphetNet & 43.68  & 20.64 & 40.72 \\
PEGASUS & 44.17 & \textbf{21.47} & \textbf{41.11} \\
\hline
ESACL & \textbf{44.24} & \textbf{21.06} & \textbf{41.20} \\
\toprule
\end{tabular}
\end{threeparttable}
\caption{ROUGE (RG) evaluation on CNN/DM dataset}
\label{table::performance-cnndm}
\end{table}

\textbf{Results on XSUM}: Table \ref{table::performance-xsum} records the ROUGE score on XSUM. ESACL outperforms the natural baseline and extractive systems. Our model achieves comparable performance to BART, and it is lower than the best-performed model PEGASUS.

\begin{table}[htbp]
\centering
\begin{threeparttable}
\begin{tabular}{lccc}
\toprule
\multicolumn{1}{c}{\textbf{Model}} & \textbf{RG-1} & \textbf{RG-2} & \textbf{RG-L} \\
\hline
Lead-1 & 16.30 & 1.60 & 11.95 \\
\hline
BERTSUM & 38.81 & 16.50 & 31.27 \\
MATCHSUM & 24.86 & 4.66 & 18.41 \\
\hline
PGNet & 29.70 & 9.21 & 23.24 \\
PGNet+Cov & 28.10 & 8.02 & 21.72 \\
BART & 45.14 & 22.27 & 37.25 \\
ProphetNet \tnote{*} & - & - & - \\
PEGASUS & \textbf{47.21} & \textbf{24.56} & \textbf{39.25} \\
\hline
ESACL & \textbf{44.64} &	\textbf{21.62} &	\textbf{36.73} \\
\toprule
\end{tabular}
\begin{tablenotes}
\footnotesize
     \item[*] ProphetNet doesn't provide the result on XSUM. 
\end{tablenotes}
\end{threeparttable}
\caption{ROUGE (RG) evaluation on XSUM dataset}
\label{table::performance-xsum}
\end{table}

The experimental results on two datasets show the effectiveness of the joint learning framework with contrastive learning, as indicated by the superior performance improvement of ESACL over many baseline models and the comparable performance to the best-performed SOTA model with much smaller architecture. Comparing to BART and PEGASUS, our model has less trainable parameters: we have a 12-layer encoder and 6-layer decoder, which is much smaller than the architecture of BART: 12-layer encoder and 12-layer decoder, and the architecture of PEGASUS: 16-layer encoder and 16-layer decoder.

\subsection{Human Evaluation}
To further examine the quality of the generated summaries by ESACL, we conduct the human evaluation. Two common indicators in the literature, \textbf{informativeness} and \textbf{fluency} are used to measure the quality of summary \cite{huang-etal-2020-knowledge, xu-etal-2020-self}. Informativeness measures whether the summary covers the important information from the input article and fluency focuses on if the generated summary is grammatically correct. We randomly select 100 articles from the XSUM test set and hire 7 fluent English speakers as our annotators to rate summaries generated by distil-BART and ESACL. They are required to give a comparison between the two generated summaries that are presented anonymously. Table \ref{table::human-evaluation} reports the human evaluation results. Overall, we find that our model is capable of capturing the key information of a document and the global semantics, which can be further demonstrated by the two example generated summaries from ESACL in Table \ref{table::example-summary}.

\begin{table}[htbp]
\centering
\begin{tabular}{l|ccc}
\toprule
 & \textbf{Win} & \textbf{Tie} & \textbf{Loss}\\
 \hline
Informativeness & 38.5\% & 24.7\% & 36.8\% \\
Fluency & 19.5\% & 61.0\% & 19.5\% \\
\hline
\toprule
\end{tabular}
\caption{Human evaluation results on XSUM dataset.}
\label{table::human-evaluation}
\end{table}

\begin{table*}
  \centering
  \begin{tabular}{ m{25em} | m{15em} } 
  \toprule   
    \multicolumn{1}{c|}{\textbf{Source article (abbreviated)}} & \multicolumn{1}{c}{\textbf{Summary by ESACL}} \\
    \hline
    The London trio are up for best UK act and best album, as well as getting two nominations in the best song category. "We got told like this morning 'Oh I think you're nominated'", said Dappy. "And I was like 'Oh yeah, which one?' And now we've got nominated for four awards. I mean, wow!" ... & N-Dubz have revealed they were surprised to be nominated for four Mobo Awards. \\
    \hline
    Since late November, Scotland's five mountain resorts have attracted 373,782 customers. The ski season is estimated to have attracted £37.5m into the local economy. With fresh snow on the slopes, CairnGorm Mountain expects skiing during the first weekend of June. Recent figures from Ski Scotland showed that this season's figures were better than the last bumper season of 2000-2001. ... & A record number of skiers and snowboarders have visited Scotland's five ski areas this winter. \\
    \toprule
  \end{tabular}
\caption{Two example summaries by ESACL on XSUM dataset.} 
\label{table::example-summary}
\end{table*}

\section{Discussion}
\subsection{Impact of Contrastive Learning Component} 
Since our model is warmed up using distil-BART, one could assume that the original distil-BART may simply need to be fine-tuned longer to achieve the same experimental results. Inspired by \citet{peinelt2020tbert}, we perform an additional experiment to finetune distil-BART using the same experimental settings. By analyzing the results in Table \ref{table::finetune-model-analysis}, we can conclude that longer finetuning does not considerably boost distil-BART's performance.

\begin{table}[htbp]
\centering
\begin{subtable}{1\linewidth}\centering
{\begin{tabular}{lccc}
\toprule
\multicolumn{1}{c}{\textbf{Model}} & \textbf{RG-1} & \textbf{RG-2} & \textbf{RG-L} \\
\hline
distil-BART & 41.23  & 19.38	 & 38.11 	 \\
ESACL & \textbf{44.24} & \textbf{21.06} & \textbf{41.20}\\
\toprule
\end{tabular}}
\caption{Performance on CNN/DM}
\label{tab:1a}
\end{subtable}
\begin{subtable}{1\linewidth}
\centering
{\begin{tabular}{lccc}
\toprule
\multicolumn{1}{c}{\textbf{Model}} & \textbf{RG-1} & \textbf{RG-2} & \textbf{RG-L} \\
\hline
distil-BART & 44.41  & 21.40	 &  36.50	 \\
ESACL & \textbf{44.64} &	\textbf{21.62} &	\textbf{36.73}\\
\toprule
\end{tabular}}
\caption{Performance on XSUM}
\label{tab:1b}
\end{subtable}
\caption{Finetune distil-BART under the same setting.} 
\label{table::finetune-model-analysis}
\end{table}

\subsection{Robustness Check}
\begin{table}[htbp]
\centering
\begin{tabular}{llcc}
\toprule
\multicolumn{2}{c}{\textbf{Metric}} & \textbf{Baseline} & \textbf{ESACL}\\
\hline
\multirow{2}{*}{Length} & Longest & 15.89 & 15.96 \\
& Shortest & 23.35 &  23.14\\
\hline
\multirow{2}{*}{Abstractive} & Most & 19.44 & 19.77 \\
& Least & 24.21 & 24.13  \\
\hline
\multirow{2}{*}{Distilled} & Most & 15.71 & 16.15 \\
& Least & 24.21 & 24.13  \\
\hline
\multirow{2}{*}{Position} & Latest & 17.30 & 17.60  \\
& Earliest & 22.22  & 22.40 \\
\toprule
\end{tabular}
\caption{Robustness Check on sub-populations defined by metrics using ROUGE-2. Baseline refers to distil-BART.} 
\label{table::robustness}
\end{table}

We perform a robustness check for ESACL to better understand the impacts of different datasets on the constrastive learning performance. Follow \citet{goel2021robustness}, we use several heuristics from literature to identify sub-populations of datasets. We first select top 10\% and bottom 10\% examples in the test set as two subpopulations based on four metrics from \citet{goel2021robustness}: \textbf{length}, \textbf{abstractiveness}, \textbf{distillation} and \textbf{position}. Then we evaluate the performance using ROUGE score on each subpopulation. Table \ref{table::robustness} shows the performance of ESACL in each population comparing to distil-BART \footnote{We report ROUGE-2 score in Table \ref{table::robustness}. We include the detailed results for other ROUGE scores in Appendix \ref{appendix::robustness-check}.}. For \textbf{length}, we find that ESACL performs better than the baseline on the longest set, which is the hardest to summarize considering a large amount of information. For the most \textbf{abstractive} set, ESACL is more capable to reconstruct the text and achieves higher performance. This is more significant on the most \textbf{distilled} set, ESACL performs much better than the baseline by improving the performance by 2.8\%. We can also identify a smaller performance gap between two sub-populations for ESACL, thereby emphasizing that ESACL performs more robustly than the baseline.

\subsection{Ablation Study}
\label{section::ablation-study}
To better understand the contribution of different modules in ESACL to the performance, we conduct an ablation study using the XSUM dataset.

\noindent \textbf{Document augmentation.} As we illustrated the importance of data augmentation in contrastive learning (see Section \ref{section::document-augmentation}), we design several document augmentations but we have not explored their impact on the summarization performance. Table \ref{table::ablation-rouge} shows the result of ESACL using different augmentation methods \footnote{we use ROUGE-2 as the evaluation metrics, and we also report the results using ROUGE-L in Appendix \ref{appendix::document-augmentation}.}. We can clearly see that (1) the performance with different combinations of augmentation in abstractive text summarization does not vary too much. (2) The augmentation method that interrupts the document structure, such as document rotation (\textbf{DR}), is usually harmful to the performance, since the structure of the input document plays an important role.
\begin{table}[htbp]
\centering

\begin{tabular}{c|c|c|c|c}
\toprule
& \textbf{RI} & \textbf{RD} & \textbf{RS} & \textbf{DR}\\
\hline
\textbf{RI} & 21.62 & 21.56 & 21.51 & 21.47 \\
\textbf{RD} & - & 21.59 & 21.58  & 21.38 \\
\textbf{RS} & - & - & 21.46 & 21.41  \\
\textbf{DR} & - & - & - & 21.11 \\
\toprule
\end{tabular}
\caption{Performance on XSUM dataset under different combinations of document augmentation.}
\label{table::ablation-rouge}
\end{table}

\noindent \textbf{Number of augmentation operations.} One question we have not answered is what is the optimal number of augmentation operations. We expect this number to be in a reasonable range: 
too large can completely change the document's structure and too small does not add enough noise. 
So we design the experiment with varied number of sentences modified in the document augmentation and Table \ref{table::ablation-different-sentences} shows the performance of ESACL under Random Deletion (\textbf{RD}) and Random Swap (\textbf{RS}) with different numbers of augmentation operations $n$. Given there are 19.77 sentences per article for XSUM on average \cite{narayan2018don}, we decide to choose $n$ from [1, 3, 5]. As we expected, both $n=5$ and $n=1$ performs worse than $n=3$. A reasonable choice for $n$ should be based on the characteristics of datasets under the guidance that data augmentation is useful to add some noises while preserving the critical information.
\begin{table}[htbp]
\centering

\begin{tabular}{c|ccc}
\toprule
 & \textbf{RG-1} & \textbf{RG-2} & \textbf{RG-L} \\
 \hline
 $n=1$ & 44.48 & 21.54 & 36.64 \\
 $n=3$ & 44.52 & 21.58 & 36.59  \\
 $n=5$ & 44.36 & 21.48 & 36.52 \\
\hline
\toprule
\end{tabular}
\caption{Performance on XSUM dataset with different numbers of augmentation operations.} 
\label{table::ablation-different-sentences}
\end{table}

\noindent \textbf{Layer freezing in the encoder.} In all the above experiments, we need to freeze some layers in the encoder because of the memory limitation. Especially for contrastive learning, it benefits from the larger batch size comparing to supervised learning \cite{chen2020simple}. This brings us to a trade-off between the batch size and the number of finetuned layers. Previous studies find that higher-level layers capture context-dependent aspects of text meaning while lower-level states model aspects of syntax \cite{peters2018deep, mou-etal-2016-transferable}. Thus, in our study, we freeze the first several $l$ layers of the encoder in ESACL. Table \ref{table::ablation-different-layers} reports the performance for different $l$'s under Random Deletion (\textbf{RD}) and Random Swap (\textbf{RS}). When $l=12$, the model is fine-tuned only using the augmented documents, which makes contrastive learning ineffective. When compared to $l=6$, we clearly see the benefit of incorporating contrastive learning into the seq2seq during fine-tuning. 
\begin{table}[htbp]
\centering
\begin{tabular}{l|ccc}
\toprule
 & \textbf{RG-1} & \textbf{RG-2} & \textbf{RG-L} \\
 \hline
 $l=6$ & 44.52 & 21.58 & 36.59 \\
 $l=9$ & 44.41 & 21.47 & 36.55  \\
 $l=12$ & 44.27 & 21.41 & 36.46 \\
\toprule
\end{tabular}
\caption{Performance on XSUM dataset when freezing the first $l$ layers in the encoder.}
\label{table::ablation-different-layers}
\end{table}

\section{Conclusion}
In this paper, we propose ESACL, an enhanced sequence-to-sequence model via contrastive learning to improve the performance of abstractive text summarization, where two critical components are jointly learned via fine-tuning. With several proposed sentence-level document augmentation, ESACL can build an autoencoder with a denoising capability through fine-tuning. We empirically evaluate ESACL on two datasets both quantitatively and qualitatively. The results demonstrate that ESACL outperforms several cutting-edge benchmarks. We also examine the impact of different augmentation strategies on the performance and explore the robustness of ESACL.

\clearpage
\bibliography{anthology,custom}
\bibliographystyle{acl_natbib}

\clearpage
\appendix
\section{Implementation}
\label{appendix::implementation}
Our implementation in this paper is warmed up using the pre-trained model from HuggingFace: for CNN/DM, we use "sshleifer/distilbart-cnn-12-6" \footnote{https://huggingface.co/sshleifer/distilbart-cnn-12-6}. For XSUM: we use "sshleifer/distilbart-xsum-12-6" \footnote{https://huggingface.co/sshleifer/distilbart-xsum-12-6}. These two models are fine-tuned on the corresponding datasets. Comparing to the original BART, distil-BART contains much less parameters, which is also able to achieve comparable performance. 

For CNN/DM, we also add the post-processing step after the generation, which is common in literature. We use the same post-processing program \footnote{https://github.com/microsoft/ProphetNet} from SOTA model ProphetNet \cite{qi2020prophetnet}.

\section{Robustness Check}
\label{appendix::robustness-check}

\begin{figure}[htbp]
\centering
\includegraphics[width=0.48\textwidth]{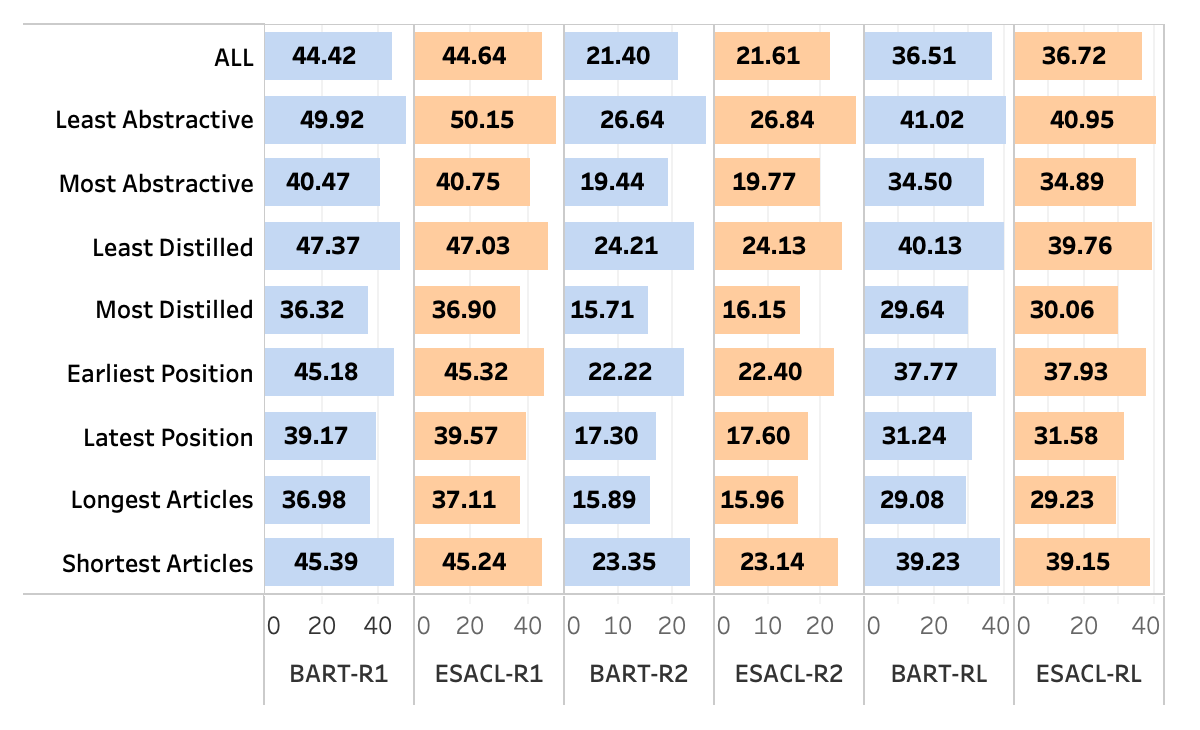}
\caption{Robustness check on XSUM dataset.}
\label{fig::robustness}
\end{figure}

For robustness check, we report the performance using ROUGE-1, ROUGE-2 and ROUGE-L in Figure \ref{fig::robustness}. We can have the same conclusion by analyzing the numbers: ESACL is more capable to perform well and robust for documents with different features, since ESACL can perform better than the baseline in the subpopulation which is more challenging to summarize. For example, for the most distilled subpopulation which requires the language model to reconstruct a lot of information from the document (refer to "Most Distilled"), ESACL performs better than the baseline by 1.6\%, 2.8\% and 1.4\% in RG-1, RG-2 and RG-L. We can also find the performance gap for each subpopulation from ESACL is smaller. The key conclusion from the robustness check is ESACL performs more stable and robust on different types of the input document.

\section{Document Augmentation}
\label{appendix::document-augmentation}

In Section \ref{section::ablation-study}, we perform an ablation study to analyze the impacts of choosing different document augmentation strategies on the model performance. Table \ref{table::ablation-rouge-2-L} records the performance evaluated by ROUGE-2 and ROUGE-L under different combinations of document augmentation methods. Numbers in parenthesis refer to the ROUGE-L score. We have the consistent conclusion with our previous analysis in Section \ref{section::ablation-study}.

\begin{table}[htbp]
\begin{tabular}{c|c|c|c|c}
\toprule
& \textbf{RI} & \textbf{RD} & \textbf{RS} & \textbf{DR}\\
\hline
\multirow{2}{*}{\textbf{RI}} & 21.62 & 21.56 & 21.51 & 21.47 \\
& (36.73) & (36.65) & (36.62) & (36.55)\\
\multirow{2}{*}{\textbf{RD}} & - & 21.59 & 21.58  & 21.38 \\
& - & (36.65) & (36.59) & (36.38)\\
\multirow{2}{*}{\textbf{RS}} & - & - & 21.46 & 21.41  \\
& - & - & (36.49) & (36.44)\\
\multirow{2}{*}{\textbf{DR}} & - & - & - & 21.11 \\
& - & - & - & (36.12) \\
\toprule
\end{tabular}
\caption{Performance on XSUM dataset under different combinations of document augmentation. We report the ROUGE-2 and ROUGE-L (in parenthesis) F1  score.}
\label{table::ablation-rouge-2-L}
\end{table}

\end{document}